\documentclass[letterpaper, 10 pt, conference]{ieeeconf}  % Comment this line out if you need a4paper

\IEEEoverridecommandlockouts                              % This command is only needed if 
\makeatletter
\let\NAT@parse\undefined
\makeatother
\usepackage[numbers,sort&compress]{natbib}

\title{\LARGE \bf
TREND: Tri-teaching for Robust Preference-based Reinforcement Learning with Demonstrations
}

\author{Shuaiyi Huang$^{1}$, Mara Levy$^{1}$, Anubhav Gupta$^{1}$, Daniel Ekpo$^{1}$, Ruijie Zheng$^{1}$, Abhinav Shrivastava$^{1}$% <-this % stops a space
\thanks{$^{1}$University of Maryland, College Park. \{huangshy, mlevy, anubhav, daniekpo, rzheng12, abhinav2\}@umd.edu}%
}

\usepackage{amsmath}
\usepackage{pifont}
\usepackage{amsfonts}
\usepackage{bbm}
\usepackage{caption}
\usepackage{subcaption}
\let\labelindent\relax
\usepackage{enumitem} % no indent for itemized
\usepackage{xspace}
\usepackage{makecell}
\usepackage{graphicx}
\usepackage{booktabs}
\usepackage{multirow}
\usepackage{tikz}
\usepackage{amsmath}
\usepackage{makecell}

\usepackage[pagebackref,breaklinks,colorlinks]{hyperref}
\usepackage[capitalize]{cleveref}
\usepackage{rotating}

\usepackage{xcolor}

\newcommand{\oursbig}{Tri-teaching for Robust Preference-based Reinforcement Learning with Demonstrations\xspace}
\newcommand{\oursshort}{TREND\xspace}

\newcommand{\mw}{Meta-world}%
\DeclareMathOperator*{\argmin}{arg\,min}

\begin{document}

\maketitle
\thispagestyle{empty}
\pagestyle{empty}

%%%%%%%%%%%%%%%%%%%%%%%%%%%%%%%%%%%%%%%%%%%%%%%%%%%%%%%%%%%%%%%%%%%%%%%%%%%%%%%%
\begin{abstract}

Preference feedback collected by human or VLM annotators is often noisy, presenting a significant challenge for preference-based reinforcement learning that relies on accurate preference labels. 
To address this challenge, we propose \oursshort, a novel framework that integrates few-shot expert demonstrations with a tri-teaching strategy for effective noise mitigation. Our method trains three reward models simultaneously, where each model views its small-loss preference pairs as useful knowledge and teaches such useful pairs to its peer network for updating the parameters. Remarkably, our approach requires as few as one to three expert demonstrations to achieve high performance. We evaluate \oursshort on various robotic manipulation tasks, achieving up to 90\% success rates even with noise levels as high as 40\%, highlighting its effective robustness in handling noisy preference feedback. Project page: \href{https://shuaiyihuang.github.io/publications/TREND}{https://shuaiyihuang.github.io/publications/TREND}.

\end{abstract}

%%%%%%%%%%%%%%%%%%%%%%%%%%%%%%%%%%%%%%%%%%%%%%%%%%%%%%%%%%%%%%%%%%%%%%%%%%%%%%%%
\section{Introduction}
A key challenge when using reinforcement learning (RL) to train autonomous agents for various tasks is defining a suitable reward function. The reward function needs to give dense feedback to guide the agent’s learning while preventing unintended behaviors (series of actions) that might earn high rewards but do not match the user's intent. 
However, creating an effective reward function is often complex and time-consuming. It involves multiple rounds of testing with different adjustments to ensure the agent learns the intended behavior without developing unplanned habits.\looseness=-1

To address these difficulties, Preference-based Reinforcement Learning (PbRL) has emerged as a promising approach. PbRL sidesteps the challenge of explicitly designing a reward function by instead using human preferences as the reward signal~\cite{lee2021pebble, park2022surf, sadigh2017active, wayex, pilarski2011online, wilson2012bayesian, wirth2016model}. 
% Preference-based Reinforcement Learning (PbRL) is a popular choice for learning a reward model. It works by leveraging human preferences as the reward signal~\cite{lee2021pebble, park2022surf, sadigh2017active, pilarski2011online, wilson2012bayesian, wirth2016model}. 
PbRL employs a human-in-the-loop methodology, where humans provide preference feedback by comparing pairs of trajectory segments and labeling which segment is closer to the goal. This approach addresses the challenge of designing explicit reward functions and can result in behaviors that are better aligned with human intent~\cite{lee2021pebble}. Recent advances in this domain have integrated vision-language models (VLMs)~\cite{wang2024rl}, which can autonomously generate preferences, potentially reducing the need for direct human involvement. 
%which segment is better given the task instruction. 

However, in practice, both human and VLM-generated preference labels can be noisy and inconsistent. Human feedback is prone to bias and VLM labels often struggle with interpreting visual content, task-specific text, and temporal dynamics. For example, if two behaviors are qualitatively similar, it is likely that the preference labels provided by either human annotators or VLMs are not meaningful.   
As shown later in our experimental analysis (Sec.~\ref{subsec:ablate}), VLM preference labels can have a noise rate up to 40\%, making them insufficient for reward learning without denoising.
To make matters worse, a recent work~\cite{lee2021b} showed that even a 10\% label corruption rate can significantly degrade performance. 
This underscores the need for robust preference-based RL methods that can effectively handle noisy labels.

\begin{figure*}[t]
  \centering
  \includegraphics[width=1.0\linewidth]{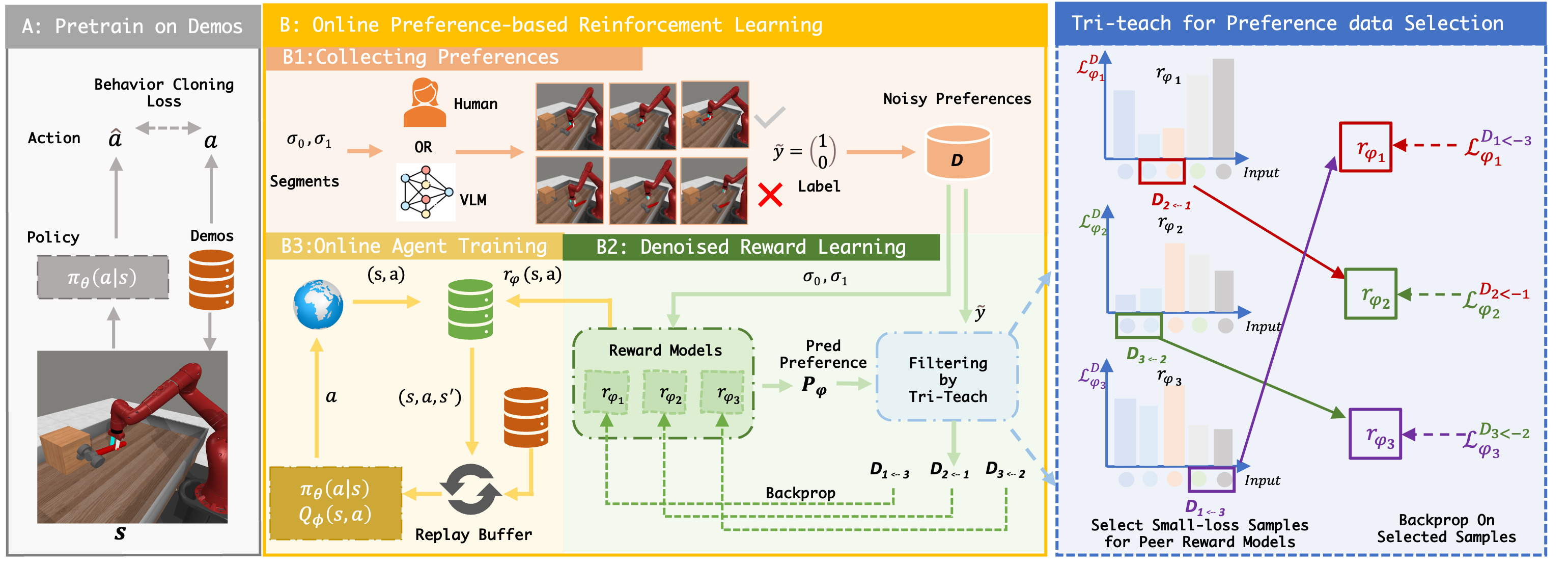}
 \caption{\small{\textbf{Overview of our method~\oursshort.} First, we pretrain the policy network using behavior cloning (BC) with few-shot expert demonstrations for effective exploration (\textcolor{gray}{A}). In the online training phase, noisy preferences are collected from human annotators or a vision-language model (\textcolor{pink}{B1}). We then apply our tri-teaching strategy for denoised reward learning, where three collaborative reward models identify clean preference samples for each other (\textcolor{lime}{B2}). Finally, the learned reward model is used to guide the agent's training (\textcolor{brown}{B3}), ensuring robust performance despite noisy labels.}}
 \label{fig:overview}
 \vspace{-0.2in}
\end{figure*}

In this work, we propose \oursbig(\oursshort), a novel framework to address the challenge of noisy preference labels, as shown in Figure~\ref{fig:overview}. 
Our key insight is the use of peer models to identify clean samples from a batch of noisy labels, instead of using a single model that can be biased in its estimation of noise. 
We therefore introduce a tri-teaching strategy where three reward models collaboratively select samples for training. 
Each model answers the question, “\textit{Which preference pair is more likely to have clean labels?}”, by computing losses over preference pairs and labeling those with small losses as likely clean. 
These selected pairs are then used to update the other models. \looseness=-1

Our tri-teaching strategy is advantageous compared with existing works that simply average multiple reward model's predictions for sample selection~\cite{cheng2024rime} for the following reasons: (a) each peer model independently develops expertise in sample selection, enhancing robustness to noise, and (b) the definition of clean samples is dynamically learned from peer models rather than being fixed.\looseness=-1  

Additionally, to further combat extremely noisy preference labels, we integrate few-shot expert demonstrations into both pretraining and online PbRL adaptation. This provides a strong initialization for policy and ensures that at least some of the training data remains free of noise during online learning. Our experimental results demonstrate that under the extremely high noise rate of 40\%, even a few demonstrations (1-3), we can achieve a success rate of ${\sim}$80\%.

We evaluate our proposed method, \oursshort, on robotic manipulation tasks from~\mw~\cite{yu2019meta}. Our approach consistently outperforms baseline methods by a large margin, even under high noise (e.g., with noise rates up to 40\%). We also conduct a detailed ablation study to demonstrate the benefits of our key components. The main contributions of this work are as follows:

\begin{itemize}
    \item We present \oursshort, a novel framework that integrates expert demonstrations for robust learning under noisy preferences.
    \item We introduce a simple yet effective tri-teaching strategy for effective label denoising, leveraging peer models for cyclic clean feedback selection.
    \item \oursshort consistently outperforms existing PbRL baselines across a range of robotic manipulation tasks on Meta-World, even under varying levels of label noise. Notably, in scenarios with an extremely high noise rate of 40\%, \oursshort achieves significant improvements in success rates: nearly 40\% on the Button-Press task, 60\% on the Drawer-Open task, and 70\% on the Hammer task, compared to baselines using the same number of demonstrations. Even with noisy VLM-generated preference labels, \oursshort demonstrates an impressive success rate improvement of over 40\% on the Drawer-Open task, showcasing its robustness in handling noisy data.\looseness=-1
\end{itemize}

\section{Related Work}
\noindent\textbf{Preference-based RL.} While providing precise rewards for many tasks can be challenging for humans, expressing preferences for certain actions or states is often more intuitive and easier. Due to the simplicity and richness of relative preferences,
% , as well as its success on large language models, 
preference-based reinforcement learning (PbRL) has gained significant attention in recent years~\cite{park2022surf, christiano2017deep, sadigh2017active, pilarski2011online, wilson2012bayesian, wirth2016model, lee2021b}. Various approaches have been proposed to enhance the sample efficiency of PbRL. For example, PEBBLE~\cite{lee2021pebble} uses unsupervised policy pre-training to warm-start the policy and encourage exploration from the outset, while~\cite{ibarz2018reward} uses imitation from expert demonstrations for policy initialization. 
Other works have focused on developing sampling strategies to select informative preference queries~\cite{sadigh2017active, biyik2018batch, biyik2020active} and addressing exploration challenges by estimating reward model uncertainty~\cite{liang2022rune}.\looseness=-1

\noindent\textbf{Noisy label learning.}
Another critical issue with PbRL is its sensitivity to noise in preference labels.
% While these advancements have significantly improved feedback efficiency, the equally critical issue of robustness in PbRL should not be overlooked. 
Lee et al.~\cite{lee2021b} showed that even a modest 10\% rate of corrupted preference labels can substantially impair algorithmic performance. This concern becomes more pronounced in broader application contexts where preferences are collected from non-experts, increasing the likelihood of noise in labels. The issue of noisy labels is well-studied in the context of supervised learning~\cite{song2022noisylabelslearning} with proposed solutions such as sample selection methods~\cite{wang2021denoising,han2018co}, architectural modifications for noise adaptation~\cite{goldberger2022noiseadaptation}, and regularization~\cite{lukasik2020does}. These methods, however, are not easily transferable to PbRL due to limited sample sizes~\cite{cheng2024rime}. To alleviate this problem in PbRL, \cite{xue2023reinforcement} introduced an encoder-decoder framework to capture diverse human preferences and improve robustness, though it required a substantially larger amount of preference data. RIME~\cite{cheng2024rime} introduced a dynamic sample selection method using a denoising discriminator to filter out the noise and incorporate a warm start strategy to mitigate errors from noisy labels. In contrast, we select reliable samples by a tri-teaching method and, optionally, explicitly incorporate a few expert demonstrations to combat extremely high noise.

\noindent\textbf{Vision Language Models (VLMs) for RL.} 
Recent advancements in integrating VLMs into RL have shown promising outcomes~\cite{zheng2024tracevla,huang2024point,huang2024uvis,choi2022lmpriors, du2023guiding,  rocamonde2023vision,zhu2017structured,huang2022learning,wang2024rl,huang2020confidence,klissarov2023motif,he2023towards,huang2019dynamic,huang2024ardup, p3po, wu2024autohallusion, wei2023imitation, zhengprise, zheng2024premier, zheng2023taco}, particularly in reward specification and preference-based methods. 
LLMs/VLMs have been employed to generate reward functions from text descriptions of goals, such as generating reward code directly~\cite{xie2024text2rewardrewardshapingllm} or serving as proxy reward functions aligned with given prompts~\cite{kwon2023reward}. 
% Other approaches have used VLMs to classify safe states, guide exploration, or generate rewards directly based on task descriptions and observations \cite{choi2022lmpriors, du2023guiding, wang2024rl, rocamonde2023vision}. 
% Beyond reward generation, LLMs/VLMs have been utilized for generating preference feedback, such as in Motif \cite{klissarov2023motif}, which uses LLM-generated preferences to create intrinsic rewards guiding the agent's exploration. 
% Extending to multimodal inputs, a recent work, RL-VLM-F \cite{wang2024rl}, uses VLMs to generate preference labels over pairs of the embodied agent's image observations. 
Beyond reward generation, VLMs have been utilized for generating preference feedback, such as in RL-VLM-F~\cite{wang2024rl}, which uses VLMs to generate preference labels over pairs of the embodied agent's image observations. 
However, a key problem with VLM-generated preference labels is that they are often unreliable leading to high noise, so directly feeding all VLM feedbacks can lead to suboptimal (or worse) performance. 
In this work, we address this issue by introducing a tri-teaching method, supplemented with a few expert demonstrations, to denoise VLM's noisy preference labels.\looseness=-1

\section{Preliminaries}

\noindent\textbf{Preference-based RL.} In classical RL~\cite{sutton2018reinforcement}, an agent learns by interacting with the environment for $T$ discrete time steps. At each step $t$ the agent observes the environment state $s_t$ and then uses a policy $\pi$ to choose an action $a_t$. After executing the action, the environment returns a reward $r$ and transitions to the next state $s_{t+1}$ following the environment's dynamics. The objective of the learning agent is to learn a policy that maximizes the expected return, which is the discounted cumulative reward over time: $R_t = \sum_{k=0}^{\infty} \gamma^k r(s_{t+k}, a_{t+k})$ where \(\gamma \in [0, 1)\) is the discount factor, which balances the agent’s preference for immediate versus long-term rewards.

Defining an explicit reward function can be difficult or infeasible in many real-world tasks. PbRL addresses this by relying on a teacher (often a human or an expert model) to provide feedback in the form of preferences between different behaviors (trajectory segments) of the agent~\cite{lee2021pebble, liu2022meta, kim2023preferencetransformer, cheng2024rime} (e.g., in Fig~\ref{fig:overview} B1). Instead of receiving scalar rewards from the environment, the agent learns from the teacher’s preferences, using them to construct an internal reward model.

Formally, a trajectory $\sigma$ is defined as a sequence of state-action pairs \( \{(s_k, a_k), \dots, (s_{k+H-1}, a_{k+H-1})\} \) over a fixed  segment size $H$. Given two trajectory segments \( \sigma_0 \) and \( \sigma_1 \), the teacher provides feedback in the form of a preference label \( \tilde{y} \in \{(1, 0), (0,1), (0.5,0.5)\} \). A ``preference'' is expressing which is preferred between pairs of clips of the agent's behavior, essentially distinguishing which segment has a higher reward. Here, \(\tilde{y}=(1,0)\) indicates \(\sigma_0 \succ \sigma_1 \). Correspondingly, \(\tilde{y}=(0,1)\) indicates \(\sigma_1 \succ \sigma_0 \), and \(\tilde{y}=(0.5,0.5)\) indicates that the teacher is indifferent between the two trajectories. The teacher's feedback is stored in a dataset \(D\) as a tuple \((\sigma_0, \sigma_1, \tilde{y})\). The agent uses this data to learn a reward function \(r_\psi\).

The probability that the teacher prefers \(\sigma_1\) over \(\sigma_0\) can be modeled using the Bradley-Terry model~\cite{bradley1952rank}, based on the estimated reward function \(r_\psi\) as
\begin{equation}
P_\psi[\sigma_i \succ \sigma_j] = \frac{\exp\left( \sum_t \hat{r}_\psi(s_t^i, a_t^i) \right)}{\sum_{i \in \{j, i\}} \exp\left( \sum_t \hat{r}_\psi(s_t^i, a_t^i) \right)}. %\tag{1}
\label{eq:p}
\end{equation}
The reward function \(r_\psi\) is learned by minimizing the cross-entropy between the predicted preference and the label provided by the teacher. The loss function is defined as:
\begin{align}
    \mathcal{L}_{\text{CE}}(\psi) = \mathbb{E}\left[\ell_{\text{Reward}}\right]
    &= -\mathbb{E}\Big[ \tilde{y}(0) \ln P_{\psi}[\sigma_0 \succ \sigma_1] \notag \\
    &\quad + \tilde{y}(1) \ln P_{\psi}[\sigma_1 \succ \sigma_0] \Big] %\tag{2}
    \label{eq:lce}
\end{align}

\section{TREND: Tri-teaching for Robust PbRL with Demonstrations}
In this section, we will describe our approach, \oursbig (\oursshort), in detail. \oursshort simultaneously trains three reward networks. Each reward model updates its parameters by using the preference pairs selected from the peer network via our proposed tri-teaching (Sec.~\ref{subsec:teach}). To further mitigate the impact of noise, we can leverage expert demonstrations in the learning process (Sec.~\ref{subsec:expert}).

\subsection{Tri-teaching for Preference Data Selection}
\label{subsec:teach}

Following the small-loss principle~\cite{han2018co}, we hypothesize that the loss of a preference pair is positively correlated with how noisy the label is. Since a single model can be biased in its estimation of noise, potentially leading to error accumulation, we introduce a tri-teaching strategy where three peer reward networks collaboratively select samples for training. Each network answers the question, “\textit{Which preference pair is more likely to have clean labels?}” and teaches the other networks by providing these samples for training. 
We use three teachers as existing work~\cite{lee2021pebble,cheng2024rime,park2022surf} used an ensemble of three reward models to improve the stability in reward learning. Using an odd number of experts~\cite{chao2024three,shazeer2017outrageously} is also a standard practice to avoid ties and provide a majority consensus signal.

Formally, we define a selection loss \( \mathcal{L}_{\psi}^{\mathcal{D}} \) to quantify how noisy the label is as estimated by the reward model \( r_{\psi} \). Given a batch of training preference data \( \mathcal{D} = \{(\sigma_0^i, \sigma_1^i, \Tilde{y}_i)\}_{i=1}^N \), where \( (\sigma_0^i, \sigma_1^i) \) is a pair of trajectory segments, and \( \Tilde{y}_i \) represents the corresponding preference label, we compute the selection loss for model \( \psi_k \) as
\begin{equation}
\mathcal{L}_{\psi_k}^{\mathcal{D}} = \frac{1}{N} \sum_{i=1}^{N} \ell_{\text{reward}}((\sigma_0^i, \sigma_1^i), \Tilde{y}_i;\psi_k), \quad k \in \{1, 2, 3\},
\label{eq:ld}
\end{equation}
where $k$ is the index of reward models.

During training, each reward model \( \psi_k \) identifies the subset of samples that have the smallest selection loss within a batch of preferences. The size of the subset is determined by a selection rate $\gamma$, which controls the proportion of samples retained. These samples are considered clean and are passed to a peer model for training. The selected data for peer training is defined as
\begin{equation}
\mathcal{D}_{k\leftarrow j} = \argmin_{\bar{\mathcal{D}}: |\bar{\mathcal{D}}| \geq \gamma|\mathcal{D}| 
\text{ and } \bar{\mathcal{D}} \subseteq \mathcal{D}} \mathcal{L}_{\psi_j}^{\bar{\mathcal{D}}}.
\label{eq:dkj}
\end{equation}

Here $(k,j) \in \{(2,1), (3,2), (1,3)\}$, and $\mathcal{D}_{k \leftarrow j}$ denotes the clean preferences identified by the reward model $r_{\psi_j}$ which are then selected as training samples for $r_{\psi_k}$. In this way, the three models exchange opinions on which samples are clean, thereby enabling each to learn from its peers.

After the selection process, each reward model updates its parameters by using the preference pairs selected from the peer network as \vspace{-0.1in}\begin{equation}
\psi_{k} = \psi_{k} - \eta \nabla_{\psi_{k}} \mathcal{L}_{\psi_k}^{D_{k\leftarrow j}}.
\label{eq:psik}
\end{equation}
This process is cyclic, where \( \hat{r}_{\psi_1} \) selects for \( \hat{r}_{\psi_2} \), \( \hat{r}_{\psi_2} \) selects for \( \hat{r}_{\psi_3} \), and \( \hat{r}_{\psi_3} \) selects for \( \hat{r}_{\psi_1} \). This tri-teaching strategy allows the models to collaboratively denoise the labels and improve learning stability despite noisy preferences.

%The selection rate \( C(t) \) dynamically adjusts the number of samples retained during training. 

\begin{figure*}[t]
  \centering
  \includegraphics[width=0.85\linewidth]{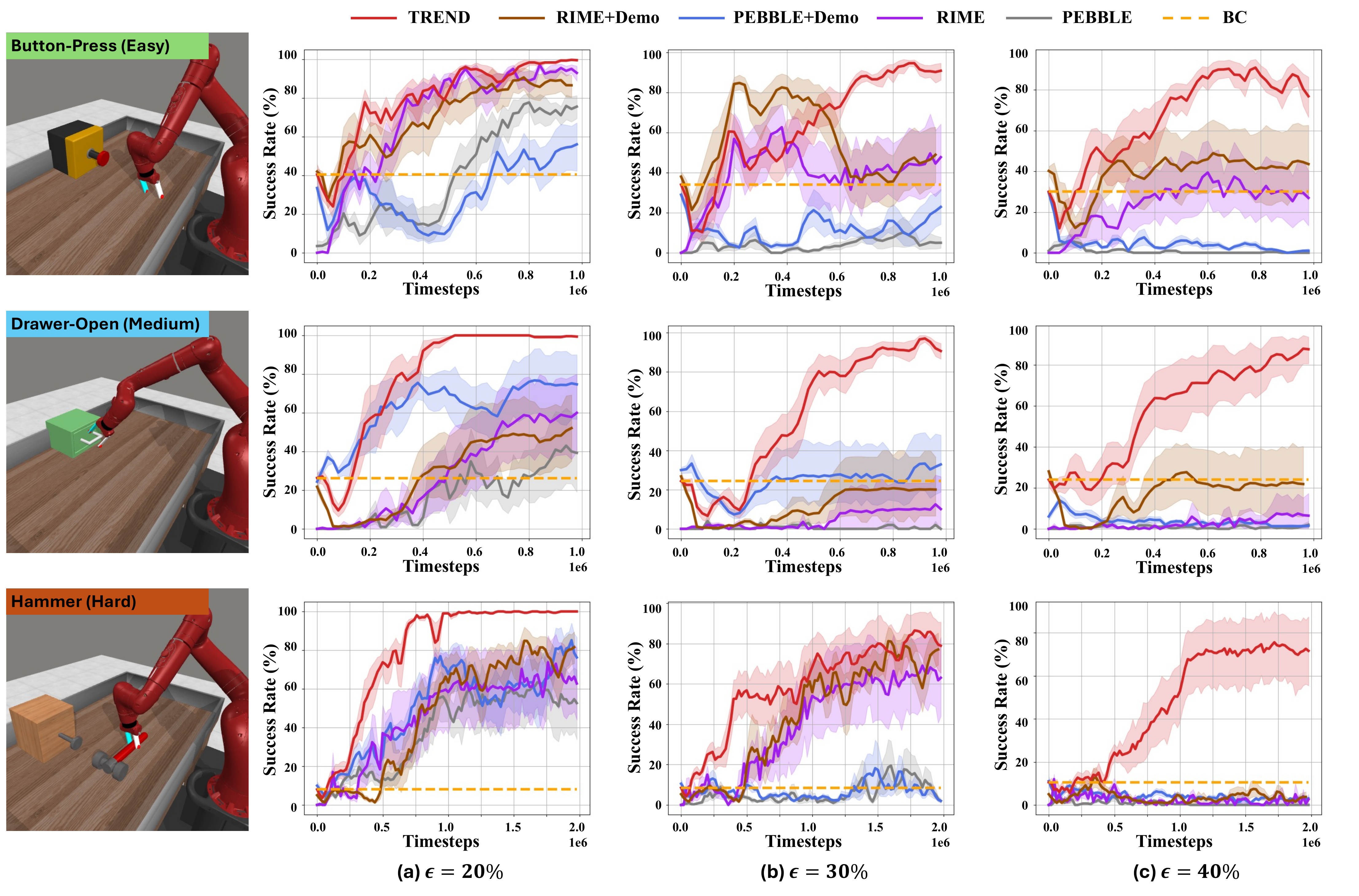}
%\vspace{-0.1in}
  \caption{\small{\textbf{Learning curves for robot manipulation tasks on \mw.} Each row represents results for a specific task and each column corresponds to a different error rate $\epsilon$. Results are averaged over five seeds. Shaded Areas represent standard deviation across seeds.}}
  \vspace{-0.2in}
 \label{fig:result_metaworld}
\end{figure*}

We ensure diversity among the reward models by using different weight initialization and permuting the input sample orders. This prevents them from collapsing into similar models, maintaining diverse learning dynamics.

\subsection{Incorporating Few-shot Expert Demonstrations}
\label{subsec:expert}

In the presence of noisy preference labels, purely learning from trial and error through online reinforcement learning can lead to sub-optimal performance. 
To further mitigate the impact of noise and to provide a clean supervisory signal, we incorporate expert demonstrations into the pertaining and online training process. 
Expert trajectories offer high-quality information, which is particularly valuable when learning under high-noise conditions. 
These demonstrations serve to guide the agent toward better behavior by ensuring that a portion of the training data is noise-free.\looseness=-1

Our method builds upon the widely used PbRL algorithm PEBBLE~\cite{lee2021pebble}. In the early stages of training, querying the teacher for preference labels often yields uninformative results, as the agent's policy tends to be random and the states uninteresting. To mitigate this issue, PEBBLE~\cite{lee2021pebble} introduced unsupervised exploration through pretraining the agent with an intrinsic reward based on state entropy~\cite{singh2003nearest}, a strategy that has been widely adopted in subsequent research~\cite{park2022surf,cheng2024rime}.

To encourage more diversified exploration, we replace the unsupervised exploration phase of the PEBBLE algorithm with policy pretraining on demonstration data using behavior cloning (BC). Even though we are only using a few demonstrations, BC pretraining is significantly more effective than PEBBLE’s intrinsic reward-based pretraining~\cite{singh2003nearest,lee2021pebble}. BC establishes a stronger prior for the policy, enhances exploration, and improves the sampling of informative preference pairs. This is especially beneficial to hard tasks where extropy-based exploration struggles to find useful signal.

Next, after the policy initialization phase by BC, during the online training phase, for each batch of training data, we sample a proportion of $\alpha$\% from the expert demonstration dataset $\mathcal{D}_{\text{expert}}$, where $\alpha$\% controls the proportion of expert demonstrations. For the remaining data, we sample from the non-expert RL replay buffer $\mathcal{D}_{\text{replay}}$. Specifically, for data from the expert demonstrations, we update the policy network $\pi_\theta$ using the BC loss as
\begin{equation}
\mathcal{L}^{\text{BC}}_{\pi_\theta} = \mathbb{E}_{(s_t, a_t) \sim \mathcal{D}_{\text{expert}}} \left[ \| \pi_\theta(s_t) - a_t \|^2 \right].
\label{eq:lbcpi}
\end{equation}
For non-expert data from the replay buffer, we apply the standard Soft Actor-Critic (SAC)~\cite{sac} loss to update the policy network $\pi_\theta$ \begin{equation*} \mathcal{L}^{\text{SAC}}_{\pi_\theta} = \mathbb{E}_{s_t \sim \mathcal{D}_{\text{replay}}, a_t\sim \pi_\theta(\cdot|s_t)} \left[ \alpha \log \pi_\theta (a_t \mid s_t) - Q_\phi (s_t, a_t) \right], \end{equation*} where $Q_\phi$ is the critic network. 
We then update the policy network $\pi_\theta$ using a weighted average of the two losses as 
\begin{equation} \mathcal{L}^{\text{\oursshort}}_{\pi_\theta} = \mathcal{L}^{\text{SAC}}_{\pi_\theta} + \lambda_{\text{BC}} \mathcal{L}^{\text{BC}}_{\pi_\theta}. 
\label{eq:ltrend}
\end{equation} 
This way, the BC loss serves as a strong regularization term during online training, helping the policy network retain its BC initialization and avoid over-reliance on noisy preference data, which is especially beneficial in high-noise conditions.

\section{Experiments}
%TODO this is copied from RIME
\subsection{Experimental Setup}

\noindent\textbf{Tasks.} We evaluate \oursshort on three robotic manipulation tasks from \mw~\cite{yu2019meta}: Button-Press, Drawer-Open, and Hammer. 
These tasks are visually rich, with diverse textures and shading, requiring precise, fine-grained control to successfully manipulate objects. The tasks are shown in the first column of Figure~\ref{fig:result_metaworld}.

To analyze our method's effectiveness, we consider two types of noisy annotators: one that generates preferences using a scripted policy and another that utilizes a VLM-based preference generator.

\noindent\textbf{Scripted Noisy Preference Annotator.} We generate synthetic preference feedback with an oracle reward following the technique used in previous works~\cite{cheng2024rime, lee2021pebble}.
Specifically, we assume a scripted teacher that determines the preference between two trajectory segments based on the sum of the ground-truth reward for each segment.
To introduce noise, we flip each preference label with a probability of $\epsilon=20\%, 30\%, 40\%$ as done in RIME~\cite{cheng2024rime}.
This allows us to easily control the level of noise in preference labels.

\noindent\textbf{VLM Preference Annotator.} We also evaluate \oursshort using preference labels generated by a VLM. This approach follows the work done in RL-VLM-F~\cite{wang2024rl}. For all VLM experiments, we use Gemini-1.5-Flash~\cite{gemini}. We prompt the VLM using rendered images of the two trajectory segments and the task description and obtain the generated preference label following~\cite{wang2024rl}. If the VLM predicts that the provided segments are too similar, it will return a ``no preference'' answer and we skip this pair for reward training~\cite{wang2024rl}.

\begin{figure}[t]
  \centering
  \includegraphics[width=0.66\linewidth]{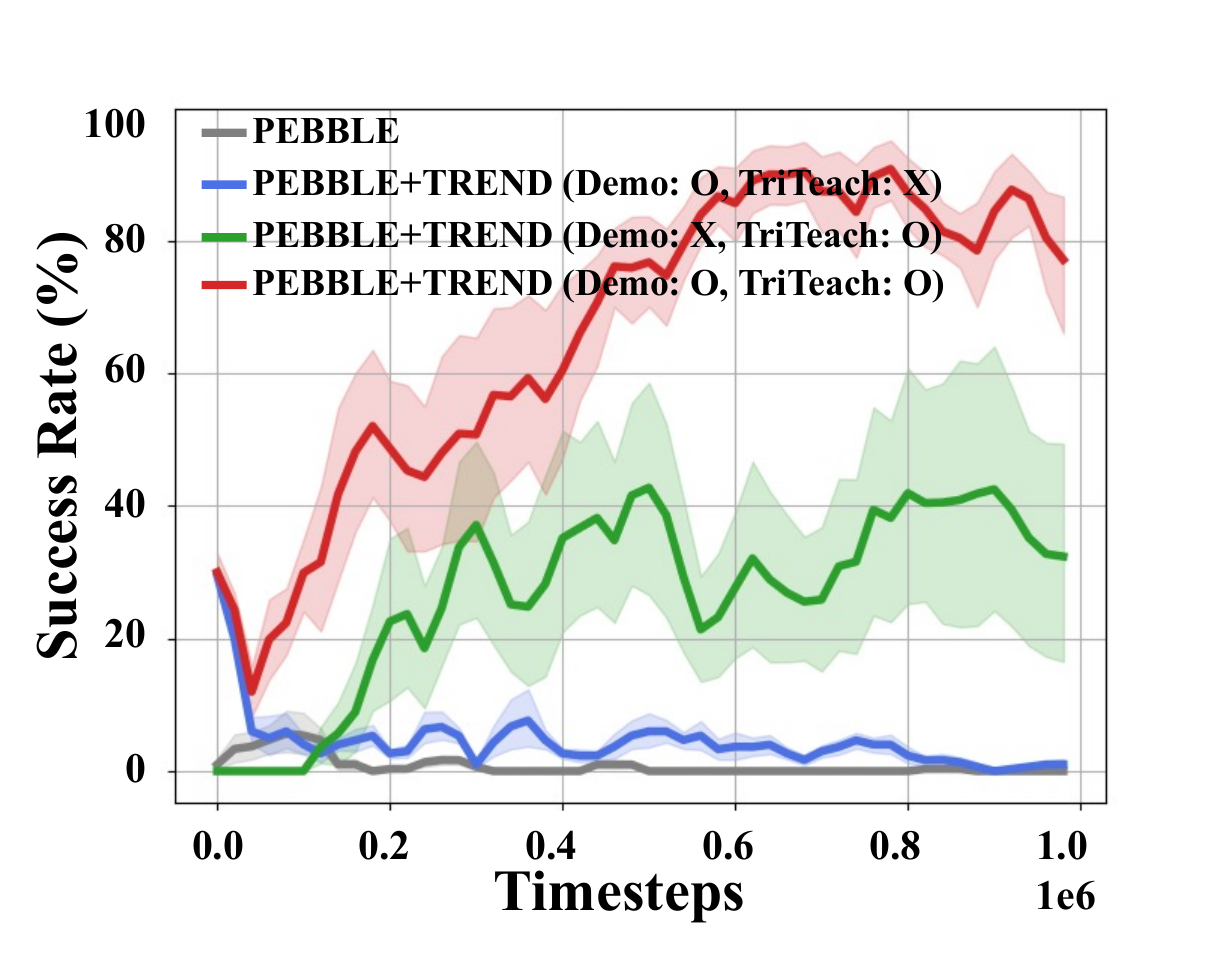}
  \vspace{-0.1in}
 \caption{\small{\textbf{Contribution of each component on Button-Press ($\epsilon=40\%$).} Our tri-teaching boosts performance significantly, and adding a single expert demonstration further enhances success, emphasizing the need for both components.}}
 \vspace{-0.18in}
 \label{fig:ablation_inv}
\end{figure}

\noindent\textbf{Implementation Details.} 
We use PEBBLE~\cite{lee2021pebble} as our backbone algorithm. For the tri-teaching part of our method, we set $\gamma = 0.6$ (in Eq.~\ref{eq:dkj}). \oursshort uses 1 expert demonstration for Button Press, 2 for Drawer Open, and 3 for Hammer, with the BC loss coefficient $\lambda = 4.0$ (in Eq.~\ref{eq:ltrend}). The proportion of sampling expert demos $\alpha\%$ is linearly decayed from 50\% to 25\% within the first 50k timesteps. We use the same hyperparameters for all experiments unless otherwise stated.

All other hyperparameters are consistent with PEBBLE. Both the baselines and our method use an ensemble of three reward models. The segment size H (i.e., length of $\sigma$) is 50. Preference queries are selected using disagreement sampling, following~\cite{cheng2024rime}. Disagreement sampling chooses pairs with high uncertainty based on the variance across an ensemble of predictors. 
All baseline methods are implemented using publicly available repositories. The initial state is randomly initialized for each episode, ensuring diverse configurations.
We run each experiment with five random seeds and report the average performance with standard deviations.

\subsection{Main Results}
Figure~\ref{fig:result_metaworld} presents the learning curves of \oursshort alongside baseline methods on~\mw, evaluated across three tasks: Button-Press (easy), Drawer-Open (moderate), and Hammer (hard). We compare our method against PEBBLE~\cite{lee2021pebble}, which lacks denoising techniques, and RIME~\cite{cheng2024rime}, the current state-of-the-art (SOTA) method for denoising in preference-based reinforcement learning (PbRL). For a fair comparison, we incorporated expert demonstrations into both PEBBLE and RIME in the same way and the same amount as ours, referring to these as PEBBLE+demo and RIME+demo. We also include a behavior cloning (BC) baseline learning purely from the demonstrations without reinforcement learning.

As shown in Figure~\ref{fig:result_metaworld}, our approach consistently outperforms all baselines across all noise levels, from 20\% to 40\%. 
Notably, we achieve over 80\% success at 40\% noise with just 1 demonstration for Button-Press, 2 for Drawer-Open, and 3 for Hammer. This success is driven by our tri-teaching strategy for label selection and the integration of expert demonstrations. On average the baselines either perform at half of our success rate or fail completely. \oursshort is also superior to PEBBLE+demo and RIME+demo, showing the benefits of \oursshort come from both the expert demonstrations and the cross-model selection, not the expert demonstrations alone. These results underscore \oursshort's robustness to noisy preferences and its effectiveness in high-noise environments.\looseness=-1

%\vspace{-1.em}
% \begin{figure}[htbp!]
\begin{figure}[t]
  \centering
  \includegraphics[width=1.0\linewidth]{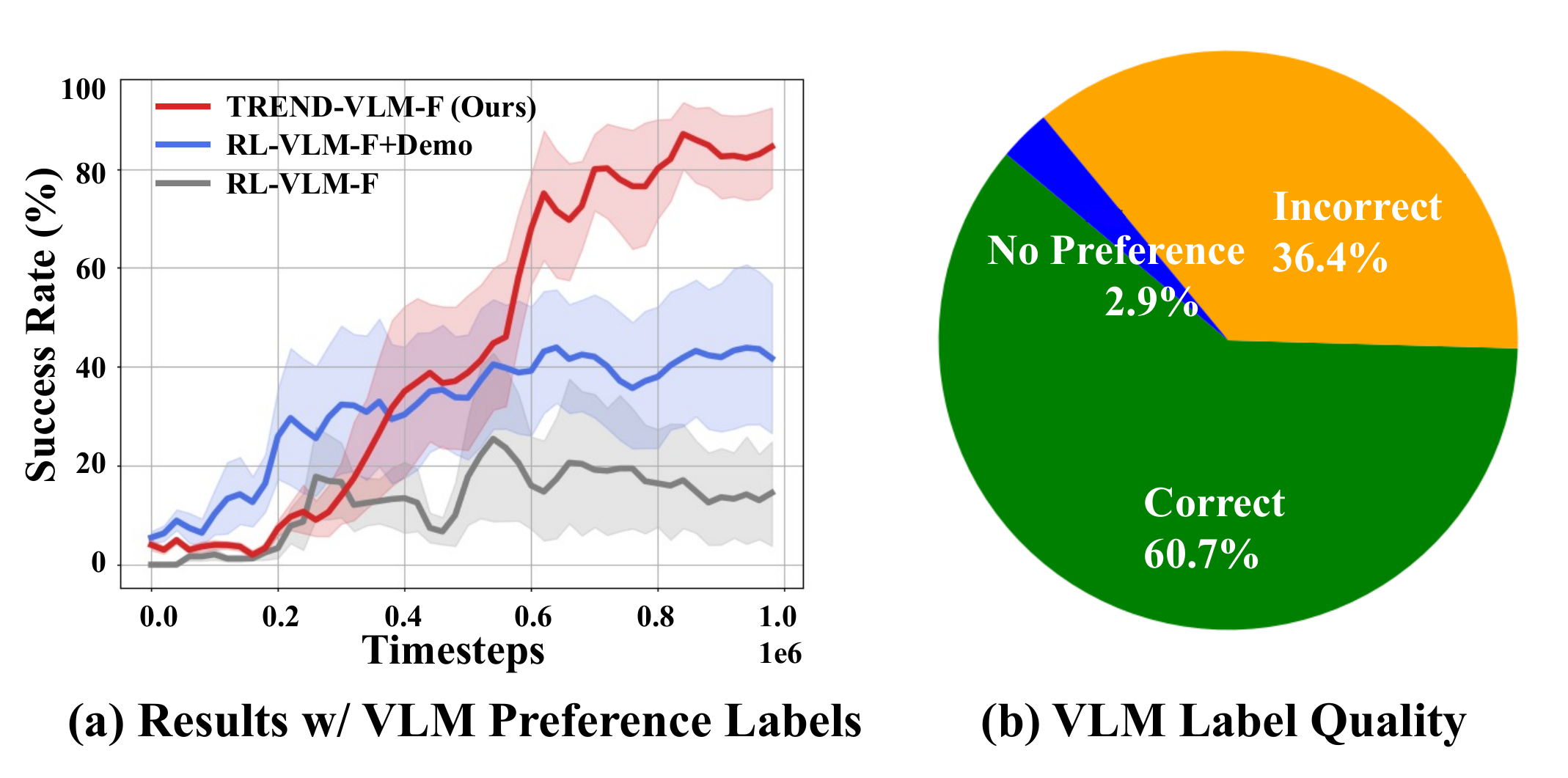}
  %\vspace{-0.2in}
 \caption{\small{\textbf{Results on Drawer-Open using VLM (Gemini-1.5-flash) to generate preference feedback.} Our \oursshort-VLM-F achives the best result (left) under the high noise rate of VLM labels (right).\looseness=-1}}
 \vspace{-0.18in}
 \label{fig:result_vlm_drawer_open}
\end{figure}
%\vspace{-0.5em}

  %\vspace{-1.em}
%\begin{figure}[!htbp]
% \begin{figure}[t]

\vspace{-1mm}
\subsection{Ablation Study}
\label{subsec:ablate}
\noindent\textbf{Effects of individual components.}
We study the effects of individual components as shown in Figure~\ref{fig:ablation_inv}. When 40\% of the labels are noisy, PEBBLE (grey) completely fails, as it lacks any denoising mechanism. However, when our tri-teaching strategy is applied, the success rate jumps to approximately 40\% (green), highlighting its effectiveness in selecting clean labels from noisy data. Adding just one expert demonstration on top of the tri-teaching (orange) boosts the success rate to around 90\%, demonstrating the critical role of even minimal clean supervision. Notably, adding an expert demo to PEBBLE alone (blue) does help avoid total failure, but is insufficient to achieve strong performance. These results underscore the necessity of both robust label selection through tri-teaching and additional clean supervision from expert demonstrations to effectively deal with high noise levels.\looseness=-1

\noindent\textbf{Performance with VLM Preference Annotator.} Figure~\ref{fig:result_vlm_drawer_open} compares our method with RL-VLM-F~\cite{wang2024rl} on the Drawer-Open task, where a VLM is used for preference labeling instead of a scripted annotator. We observe that even with an advanced proprietary VLM such as  Gemini-1.5-Flash, the generated preference labels still have a noise rate as high as 36.4\%. Consequently, as shown in the figure, RL-VLM-F (grey) achieves only a ${\sim}$20\% success rate due to the high noise level, although adding one expert demonstration improves its success rate to about 40\% (blue).\footnote{Note that the results reported in the original RL-VLM-F paper were obtained using a deterministic environment based on its released codebase, whereas we follow standard practice by randomizing the initial robot arm position and goal location.} 
In contrast, our method, \oursshort-VLM-F, achieves the best performance, reaching a ${\sim}$90\% average success rate. This demonstrates \oursshort's effectiveness in handling high noise VLM-generated preference labels.

We also tested on Hammer and Button Press, but the limitations of the current VLM preference annotator, Gemini-1.5-Flash, led to high noise rates of up to 50\%, causing both the baseline and our method to struggle. These results underscore the challenges that current VLMs face, such as difficulties in adapting to the visual appearance of simulations, understanding visual and temporal content, and selecting optimal viewpoints. 
While future advances in VLM development may help mitigate these issues, our efforts for denoising preference labels address a complementary aspect of improving VLM preference annotation quality. %\mara{is this paragraph necessary?}

\begin{figure}[t]
  \centering
  \includegraphics[width=1.0\linewidth]{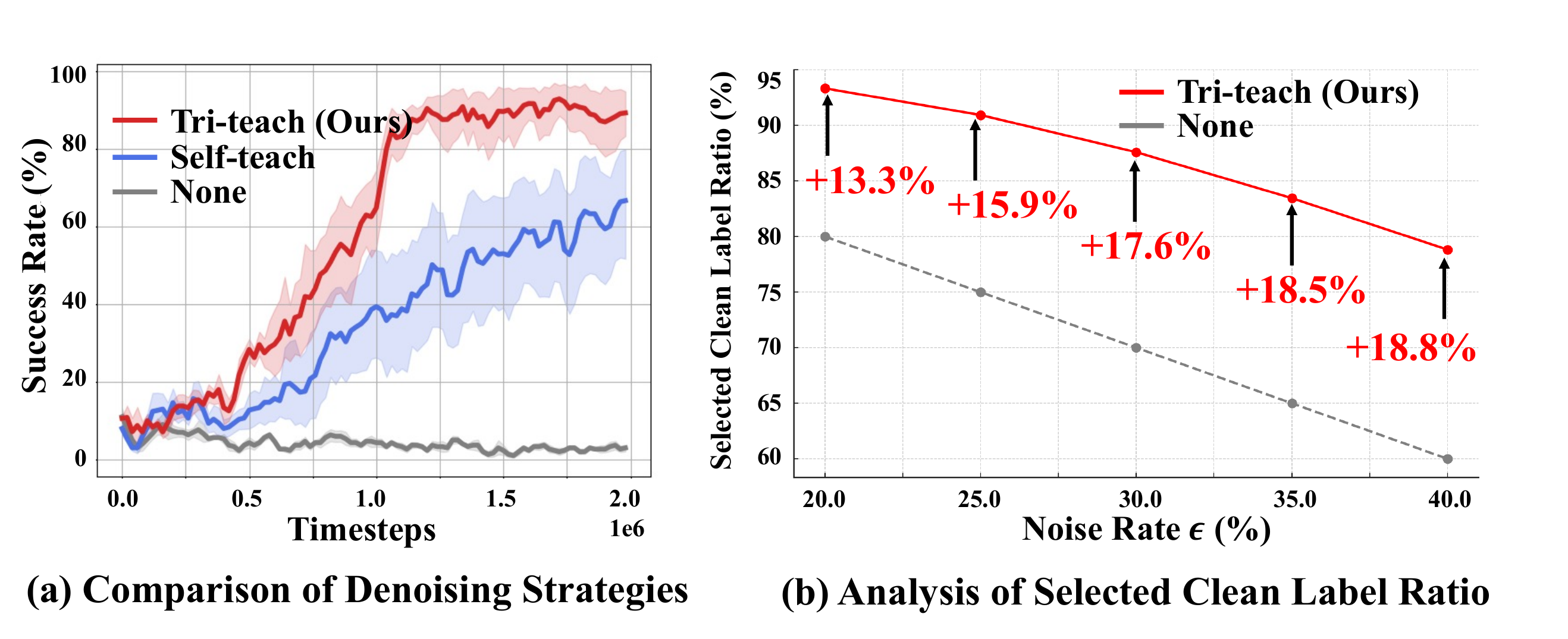}
  %\vspace{-0.2in}
     \caption{\small{\textbf{(Left):} Comparison between our method (w. Tri-teach) and a baseline denoising strategy (w. Self-teach) on Hammer ($\epsilon=40\%$). \textbf{(Right):} Comparison of clean label ratio under different noise levels between our method (w. Tri-teach) and the baseline (w/o. Tri-teach) on Hammer.}}
 \vspace{-0.2in}
 \label{fig:ablation_select}
\end{figure}

\begin{figure}[t]
  \centering
  \includegraphics[width=1.0\linewidth]{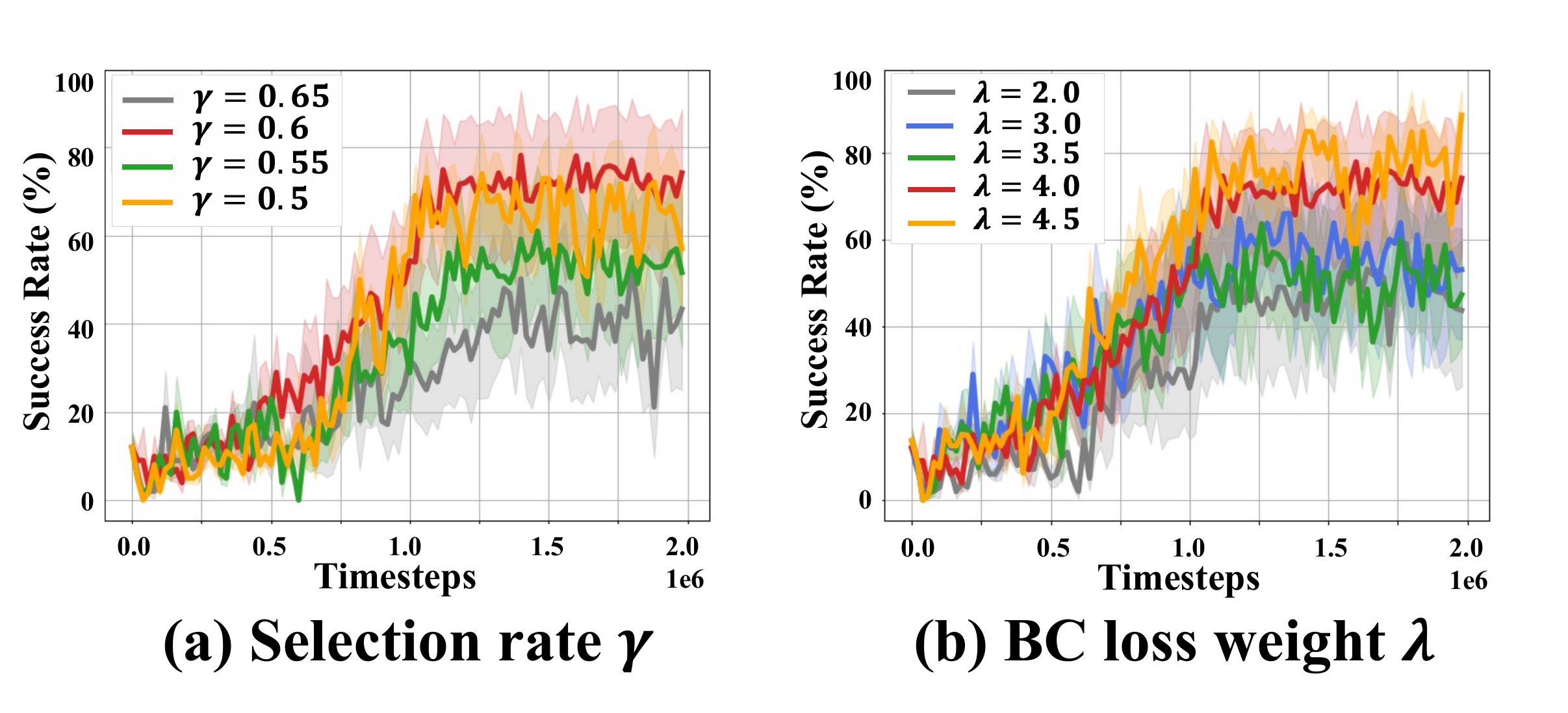}
  \vspace{-0.25in}
 \caption{\small{\textbf{Hyperparameter analysis on Hammer} ($\epsilon$=40\%).}}
 \vspace{-0.2in}
 \label{fig:ablate_hyper}
\end{figure}

\noindent\textbf{Is cyclic preference data selection and training really necessary?}
In Figure~\ref{fig:ablation_select} (left), we compare~\oursshort with a baseline denoising strategy on the Hammer task with $\epsilon=40\%$. 
Instead of cyclically selecting data with a small loss and passing it to another model for training, the baseline performs self-teaching, where each model uses its own small-loss samples for training. 
As shown in the figure, our method achieves the highest success rate, while the ablated self-teach approach performs worse than ours but still better than PEBBLE, which does not have any denoising mechanism. 
This demonstrates that selecting samples based on small loss can reduce noise to some extent, and that cyclic cross-model selection can further help denoising the preference labels.

\noindent\textbf{Analysis of label quality with Tri-teach.}
In Figure~\ref{fig:ablation_select} (right), we further analyze the clean preference label ratio achieved by our tri-teaching denoising method. 
Here, clean preference label ratio refers to the proportion of training data that has correct preference labels. On the Hammer task, using a scripted preference annotator, we documented the clean label ratio under varying noise levels (20\%, 25\%, 30\%, 35\%, and 40\%) and compared it to the baseline PEBBLE without any denoising mechanisms. 
As shown in the figure, with our tri-teach mechanism, even at a noise ratio as high as 40\%, the clean label ratio reaches ${\sim}$80\%. 
This demonstrates that our method significantly enhances preference label quality, particularly in high noise conditions.

\noindent\textbf{Impact of hyperparameters.} We analyze how the hyperparameters of \oursshort influence the performance of PbRL under a high noise rate of 40\%. In Figure~\ref{fig:ablate_hyper}, we present the learning curves for \oursshort with varying hyperparameters: (a) selection rate $\gamma \in \{50\%, 55\%, 60\%, 65\%\}$ and (b) BC loss weight $\lambda \in \{2.0, 3.0, 3.5, 4.0, 4.5\}$. We observe that \oursshort is generally robust to changes in $\gamma$, but performance can degrade when $\gamma$ is too high or too low. 
This is likely due to selecting more noisy labels or insufficient data. Ideally, selecting a $\gamma$ that aligns with the estimated clean label rate can optimize performance. Regarding the BC loss weight $\lambda$, \oursshort shows robustness across different values, with potential performance boosts when appropriately tuned.

%\vspace{-1mm}
\section{Conclusion}
In this work, we study learning from noisy preference labels in PbRL formulation. We propose \oursshort, which leverages a tri-teaching strategy and incorporates few-shot expert demonstrations. 
Through cyclic peer training among three reward networks, \oursshort effectively filters out incorrect preference labels, improving label quality even under high noise levels. 
The integration of a few expert demonstrations further enhances its performance, providing a strong supervisory signal and regularization. 
Our experimental results on robotic manipulation tasks on \mw~demonstrate that \oursshort consistently outperforms baseline methods, achieving high success rates even when the noise in the preference labels reaches up to 40\%, highlighting the robustness of our methods under noisy preference labels.

\noindent\small{\textbf{Acknowledgement} We acknowledge the constructive suggestions from reviewers and Eric Zhu. This work was partially supported by NSF CAREER Award (\#2238769) to AS. The U.S. Government is authorized to reproduce and distribute reprints for Governmental purposes notwithstanding any copyright annotation thereon. The views and conclusions contained herein are those of the authors and should not be interpreted as necessarily representing the official policies or endorsements, either expressed or implied, of NSF or the U.S. Government.}

\newpage
\bibliographystyle{IEEEtran}
\bibliography{pbrl_2024}

\begin{thebibliography}{10}
\providecommand{\url}[1]{#1}
\csname url@rmstyle\endcsname
\providecommand{\newblock}{\relax}
\providecommand{\bibinfo}[2]{#2}
\providecommand\BIBentrySTDinterwordspacing{\spaceskip=0pt\relax}
\providecommand\BIBentryALTinterwordstretchfactor{4}
\providecommand\BIBentryALTinterwordspacing{\spaceskip=\fontdimen2\font plus
\BIBentryALTinterwordstretchfactor\fontdimen3\font minus \fontdimen4\font\relax}
\providecommand\BIBforeignlanguage[2]{{%
\expandafter\ifx\csname l@#1\endcsname\relax
\typeout{** WARNING: IEEEtran.bst: No hyphenation pattern has been}%
\typeout{** loaded for the language `#1'. Using the pattern for}%
\typeout{** the default language instead.}%
\else
\language=\csname l@#1\endcsname
\fi
#2}}

\bibitem{lee2021pebble}
K.~Lee, L.~Smith, and P.~Abbeel, ``Pebble: Feedback-efficient interactive reinforcement learning via relabeling experience and unsupervised pre-training,'' in \emph{Proceedings of the 38th International Conference on Machine Learning (ICML)}, 2021.

\bibitem{park2022surf}
J.~Park, Y.~Seo, J.~Shin, H.~Lee, P.~Abbeel, and K.~Lee, ``Surf: Semi-supervised reward learning with data augmentation for feedback-efficient preference-based reinforcement learning,'' in \emph{International Conference on Learning Representations (ICLR)}, 2022.

\bibitem{sadigh2017active}
\BIBentryALTinterwordspacing
D.~Sadigh, A.~D. Dragan, S.~S. Sastry, and S.~A. Seshia, ``Active preference-based learning of reward functions,'' in \emph{Robotics: Science and Systems}, 2017. [Online]. Available: \url{https://api.semanticscholar.org/CorpusID:12226563}
\BIBentrySTDinterwordspacing

\bibitem{wayex}
M.~Levy, N.~Saini, and A.~Shrivastava, ``Wayex: Waypoint exploration using a single demonstration,'' in \emph{International Conference on Robotics and Automaction(ICRA)}, 2024.

\bibitem{pilarski2011online}
P.~M. Pilarski, M.~R. Dawson, T.~Degris, F.~Fahimi, J.~P. Carey, and R.~S. Sutton, ``Online human training of a myoelectric prosthesis controller via actor-critic reinforcement learning,'' in \emph{2011 IEEE international conference on rehabilitation robotics}.\hskip 1em plus 0.5em minus 0.4em\relax IEEE, 2011, pp. 1--7.

\bibitem{wilson2012bayesian}
A.~Wilson, A.~Fern, and P.~Tadepalli, ``A bayesian approach for policy learning from trajectory preference queries,'' \emph{Advances in neural information processing systems}, vol.~25, 2012.

\bibitem{wirth2016model}
C.~Wirth, J.~F{\"u}rnkranz, and G.~Neumann, ``Model-free preference-based reinforcement learning,'' in \emph{Proceedings of the AAAI conference on artificial intelligence}, vol.~30, no.~1, 2016.

\bibitem{wang2024rl}
Y.~Wang, Z.~Sun, J.~Zhang, Z.~Xian, E.~Biyik, D.~Held, and Z.~Erickson, ``Rl-vlm-f: Reinforcement learning from vision language foundation model feedback,'' \emph{arXiv preprint arXiv:2402.03681}, 2024.

\bibitem{lee2021b}
K.~Lee, L.~Smith, A.~Dragan, and P.~Abbeel, ``B-pref: Benchmarking preference-based reinforcement learning,'' in \emph{Proceedings of the Neural Information Processing Systems (NeurIPS) Datasets and Benchmarks Track}, 2021.

\bibitem{cheng2024rime}
J.~Cheng, G.~Xiong, X.~Dai, Q.~Miao, Y.~Lv, and F.-Y. Wang, ``Rime: Robust preference-based reinforcement learning with noisy preferences,'' in \emph{Proceedings of the 41st International Conference on Machine Learning (ICML)}, 2024.

\bibitem{yu2019meta}
\BIBentryALTinterwordspacing
T.~Yu, D.~Quillen, Z.~He, R.~Julian, K.~Hausman, C.~Finn, and S.~Levine, ``Meta-world: A benchmark and evaluation for multi-task and meta reinforcement learning,'' in \emph{Conference on Robot Learning (CoRL)}, 2019. [Online]. Available: \url{https://arxiv.org/abs/1910.10897}
\BIBentrySTDinterwordspacing

\bibitem{christiano2017deep}
P.~F. Christiano, J.~Leike, T.~Brown, M.~Martic, S.~Legg, and D.~Amodei, ``Deep reinforcement learning from human preferences,'' \emph{Advances in neural information processing systems}, vol.~30, 2017.

\bibitem{ibarz2018reward}
B.~Ibarz, J.~Leike, T.~Pohlen, G.~Irving, S.~Legg, and D.~Amodei, ``Reward learning from human preferences and demonstrations in atari,'' \emph{Advances in neural information processing systems}, vol.~31, 2018.

\bibitem{biyik2018batch}
E.~Biyik and D.~Sadigh, ``Batch active preference-based learning of reward functions,'' in \emph{Conference on robot learning}.\hskip 1em plus 0.5em minus 0.4em\relax PMLR, 2018, pp. 519--528.

\bibitem{biyik2020active}
E.~B{\i}y{\i}k, N.~Huynh, M.~J. Kochenderfer, and D.~Sadigh, ``Active preference-based gaussian process regression for reward learning,'' \emph{arXiv preprint arXiv:2005.02575}, 2020.

\bibitem{liang2022rune}
\BIBentryALTinterwordspacing
X.~Liang, K.~Shu, K.~Lee, and P.~Abbeel, ``Reward uncertainty for exploration in preference-based reinforcement learning,'' 2022. [Online]. Available: \url{https://arxiv.org/abs/2205.12401}
\BIBentrySTDinterwordspacing

\bibitem{song2022noisylabelslearning}
H.~Song, M.~Kim, D.~Park, Y.~Shin, and J.-G. Lee, ``Learning from noisy labels with deep neural networks: A survey,'' \emph{IEEE transactions on neural networks and learning systems}, vol.~34, no.~11, pp. 8135--8153, 2022.

\bibitem{wang2021denoising}
W.~Wang, F.~Feng, X.~He, L.~Nie, and T.-S. Chua, ``Denoising implicit feedback for recommendation,'' in \emph{Proceedings of the 14th ACM international conference on web search and data mining}, 2021, pp. 373--381.

\bibitem{han2018co}
B.~Han, Q.~Yao, X.~Yu, G.~Niu, M.~Xu, W.~Hu, I.~Tsang, and M.~Sugiyama, ``Co-teaching: Robust training of deep neural networks with extremely noisy labels,'' \emph{Advances in neural information processing systems}, vol.~31, 2018.

\bibitem{goldberger2022noiseadaptation}
J.~Goldberger and E.~Ben-Reuven, ``Training deep neural-networks using a noise adaptation layer,'' in \emph{International conference on learning representations}, 2022.

\bibitem{lukasik2020does}
M.~Lukasik, S.~Bhojanapalli, A.~Menon, and S.~Kumar, ``Does label smoothing mitigate label noise?'' in \emph{International Conference on Machine Learning}.\hskip 1em plus 0.5em minus 0.4em\relax PMLR, 2020, pp. 6448--6458.

\bibitem{xue2023reinforcement}
W.~Xue, B.~An, S.~Yan, and Z.~Xu, ``Reinforcement learning from diverse human preferences,'' \emph{arXiv preprint arXiv:2301.11774}, 2023.

\bibitem{zheng2024tracevla}
R.~Zheng, Y.~Liang, S.~Huang, J.~Gao, H.~Daum{\'e}~III, A.~Kolobov, F.~Huang, and J.~Yang, ``Tracevla: Visual trace prompting enhances spatial-temporal awareness for generalist robotic policies,'' \emph{arXiv preprint arXiv:2412.10345}, 2024.

\bibitem{huang2024point}
S.~Huang, D.-A. Huang, Z.~Yu, S.~Lan, S.~Radhakrishnan, J.~M. Alvarez, A.~Shrivastava, and A.~Anandkumar, ``What is point supervision worth in video instance segmentation?'' in \emph{Proceedings of the IEEE/CVF Conference on Computer Vision and Pattern Recognition Workshops (CVPRW)}, 2024, pp. 2671--2681.

\bibitem{huang2024uvis}
S.~Huang, S.~Suri, K.~Gupta, S.~S. Rambhatla, S.-n. Lim, and A.~Shrivastava, ``Uvis: Unsupervised video instance segmentation,'' in \emph{Proceedings of the IEEE/CVF Conference on Computer Vision and Pattern Recognition Workshops (CVPRW)}, 2024, pp. 2682--2692.

\bibitem{choi2022lmpriors}
K.~Choi, C.~Cundy, S.~Srivastava, and S.~Ermon, ``Lmpriors: Pre-trained language models as task-specific priors,'' \emph{arXiv preprint arXiv:2210.12530}, 2022.

\bibitem{du2023guiding}
Y.~Du, O.~Watkins, Z.~Wang, C.~Colas, T.~Darrell, P.~Abbeel, A.~Gupta, and J.~Andreas, ``Guiding pretraining in reinforcement learning with large language models,'' in \emph{International Conference on Machine Learning}.\hskip 1em plus 0.5em minus 0.4em\relax PMLR, 2023, pp. 8657--8677.

\bibitem{rocamonde2023vision}
J.~Rocamonde, V.~Montesinos, E.~Nava, E.~Perez, and D.~Lindner, ``Vision-language models are zero-shot reward models for reinforcement learning,'' \emph{arXiv preprint arXiv:2310.12921}, 2023.

\bibitem{zhu2017structured}
C.~Zhu, Y.~Zhao, S.~Huang, K.~Tu, and Y.~Ma, ``Structured attentions for visual question answering,'' in \emph{Proceedings of the IEEE International Conference on Computer Vision}, 2017, pp. 1291--1300.

\bibitem{huang2022learning}
S.~Huang, L.~Yang, B.~He, S.~Zhang, X.~He, and A.~Shrivastava, ``Learning semantic correspondence with sparse annotations,'' in \emph{European Conference on Computer Vision}.\hskip 1em plus 0.5em minus 0.4em\relax Springer, 2022, pp. 267--284.

\bibitem{huang2020confidence}
S.~Huang, Q.~Wang, and X.~He, ``Confidence-aware adversarial learning for self-supervised semantic matching,'' in \emph{Chinese Conference on Pattern Recognition and Computer Vision (PRCV)}.\hskip 1em plus 0.5em minus 0.4em\relax Springer, 2020, pp. 91--103.

\bibitem{klissarov2023motif}
\BIBentryALTinterwordspacing
M.~Klissarov, P.~D'Oro, S.~Sodhani, R.~Raileanu, P.-L. Bacon, P.~Vincent, A.~Zhang, and M.~Henaff, ``Motif: Intrinsic motivation from artificial intelligence feedback,'' 2023. [Online]. Available: \url{https://arxiv.org/abs/2310.00166}
\BIBentrySTDinterwordspacing

\bibitem{he2023towards}
B.~He, X.~Yang, H.~Wang, Z.~Wu, H.~Chen, S.~Huang, Y.~Ren, S.-N. Lim, and A.~Shrivastava, ``Towards scalable neural representation for diverse videos,'' in \emph{Proceedings of the IEEE/CVF Conference on Computer Vision and Pattern Recognition}, 2023, pp. 6132--6142.

\bibitem{huang2019dynamic}
S.~Huang, Q.~Wang, S.~Zhang, S.~Yan, and X.~He, ``Dynamic context correspondence network for semantic alignment,'' in \emph{Proceedings of the IEEE/CVF International Conference on Computer Vision}, 2019, pp. 2010--2019.

\bibitem{huang2024ardup}
S.~Huang, M.~Levy, Z.~Jiang, A.~Anandkumar, Y.~Zhu, L.~Fan, D.-A. Huang, and A.~Shrivastava, ``Ardup: Active region video diffusion for universal policies,'' \emph{arXiv preprint arXiv:2406.13301}, 2024.

\bibitem{p3po}
M.~Levy, S.~Haldar, L.~Pinto, and A.~Shirivastava, ``P3-po: Prescriptive point priors for visuo-spatial generalization of robot policies,'' in \emph{International Conference on Robotics and Automaction(ICRA)}, 2024.

\bibitem{wu2024autohallusion}
X.~Wu, T.~Guan, D.~Li, S.~Huang, X.~Liu, X.~Wang, R.~Xian, A.~Shrivastava, F.~Huang, J.~L. Boyd-Graber, \emph{et~al.}, ``Autohallusion: Automatic generation of hallucination benchmarks for vision-language models,'' \emph{arXiv preprint arXiv:2406.10900}, 2024.

\bibitem{wei2023imitation}
Y.~Wei, Y.~Sun, R.~Zheng, S.~Vemprala, R.~Bonatti, S.~Chen, R.~Madaan, Z.~Ba, A.~Kapoor, and S.~Ma, ``Is imitation all you need? generalized decision-making with dual-phase training,'' \emph{arXiv preprint arXiv:2307.07909}, 2023.

\bibitem{zhengprise}
\BIBentryALTinterwordspacing
R.~Zheng, C.-A. Cheng, H.~D. III, F.~Huang, and A.~Kolobov, ``{PRISE}: {LLM}-style sequence compression for learning temporal action abstractions in control,'' in \emph{Forty-first International Conference on Machine Learning}, 2024. [Online]. Available: \url{https://openreview.net/forum?id=p225Od0aYt}
\BIBentrySTDinterwordspacing

\bibitem{zheng2024premier}
R.~Zheng, Y.~Liang, X.~Wang, S.~Ma, H.~Daum{\'e}~III, H.~Xu, J.~Langford, P.~Palanisamy, K.~S. Basu, and F.~Huang, ``Premier-taco is a few-shot policy learner: Pretraining multitask representation via temporal action-driven contrastive loss,'' in \emph{Forty-first International Conference on Machine Learning}, 2024.

\bibitem{zheng2023taco}
\BIBentryALTinterwordspacing
R.~Zheng, X.~Wang, Y.~Sun, S.~Ma, J.~Zhao, H.~Xu, H.~D. III, and F.~Huang, ``\texttt{TACO}: Temporal latent action-driven contrastive loss for visual reinforcement learning,'' in \emph{Thirty-seventh Conference on Neural Information Processing Systems}, 2023. [Online]. Available: \url{https://openreview.net/forum?id=ezCsMOy1w9}
\BIBentrySTDinterwordspacing

\bibitem{xie2024text2rewardrewardshapingllm}
\BIBentryALTinterwordspacing
T.~Xie, S.~Zhao, C.~H. Wu, Y.~Liu, Q.~Luo, V.~Zhong, Y.~Yang, and T.~Yu, ``Text2reward: Reward shaping with language models for reinforcement learning,'' 2024. [Online]. Available: \url{https://arxiv.org/abs/2309.11489}
\BIBentrySTDinterwordspacing

\bibitem{kwon2023reward}
M.~Kwon, S.~M. Xie, K.~Bullard, and D.~Sadigh, ``Reward design with language models,'' \emph{arXiv preprint arXiv:2303.00001}, 2023.

\bibitem{sutton2018reinforcement}
R.~S. Sutton, ``Reinforcement learning: An introduction,'' \emph{A Bradford Book}, 2018.

\bibitem{liu2022meta}
R.~Liu, F.~Bai, Y.~Du, and Y.~Yang, ``Meta-reward-net: Implicitly differentiable reward learning for preference-based reinforcement learning,'' \emph{Advances in Neural Information Processing Systems}, vol.~35, pp. 22\,270--22\,284, 2022.

\bibitem{kim2023preferencetransformer}
C.~Kim, J.~Park, J.~Shin, H.~Lee, P.~Abbeel, and K.~Lee, ``Preference transformer: Modeling human preferences using transformers for rl,'' \emph{arXiv preprint arXiv:2303.00957}, 2023.

\bibitem{bradley1952rank}
R.~A. Bradley and M.~E. Terry, ``Rank analysis of incomplete block designs: I. the method of paired comparisons,'' \emph{Biometrika}, vol.~39, no. 3/4, pp. 324--345, 1952.

\bibitem{chao2024three}
G.~Chao, K.~Zhang, X.~Wang, and D.~Chu, ``Three-teaching: A three-way decision framework to handle noisy labels,'' \emph{Applied Soft Computing}, vol. 154, p. 111400, 2024.

\bibitem{shazeer2017outrageously}
N.~Shazeer, A.~Mirhoseini, K.~Maziarz, A.~Davis, Q.~Le, G.~Hinton, and J.~Dean, ``Outrageously large neural networks: The sparsely-gated mixture-of-experts layer,'' \emph{arXiv preprint arXiv:1701.06538}, 2017.

\bibitem{singh2003nearest}
H.~Singh, N.~Misra, V.~Hnizdo, A.~Fedorowicz, and E.~Demchuk, ``Nearest neighbor estimates of entropy,'' \emph{American journal of mathematical and management sciences}, vol.~23, no. 3-4, pp. 301--321, 2003.

\bibitem{sac}
T.~Haarnoja, A.~Zhou, P.~Abbeel, and S.~Levine, ``Soft actor-critic: Off-policy maximum entropy deep reinforcement learning with a stochastic actor,'' in \emph{Proceedings of the 35th International Conference on Machine Learning}, ser. Proceedings of Machine Learning Research, J.~Dy and A.~Krause, Eds., vol.~80.\hskip 1em plus 0.5em minus 0.4em\relax PMLR, 10--15 Jul 2018, pp. 1861--1870.

\bibitem{gemini}
\BIBentryALTinterwordspacing
G.~Team, ``Gemini: A family of highly capable multimodal models,'' 2024. [Online]. Available: \url{https://arxiv.org/abs/2312.11805}
\BIBentrySTDinterwordspacing

\end{thebibliography}

\end{document}